\documentclass[10pt,twocolumn,letterpaper]{article}

\usepackage[pagenumbers]{cvpr} 

\usepackage[dvipsnames]{xcolor}
\usepackage{pifont}
\usepackage{booktabs}
\usepackage[ruled,vlined]{algorithm}
\definecolor{commentcolor}{RGB}{110,154,155}   

\newcommand{\decr}[1]{{{\scriptsize{(-#1)}}}}
\newcommand{\incr}[1]{{{\scriptsize{(+#1)}}}}

\makeatletter
\def\@fnsymbol#1{\ensuremath{\ifcase#1\or \dagger\or \ddagger\or
\mathsection\or \mathparagraph\or \|\or **\or \dagger\dagger
\or \ddagger\ddagger \else\@ctrerr\fi}}
\makeatother

\definecolor{cvprblue}{rgb}{0.21,0.49,0.74}
\usepackage[pagebackref,breaklinks,colorlinks,citecolor=cvprblue]{hyperref}

\usepackage{soul}

\usepackage{textcomp}
\def\paperID{13039} 
\def\confName{CVPR}
\def\confYear{2024}

\title{On the Robustness of Large Multimodal Models Against Image Adversarial Attacks}

\author{
Xuanming Cui\thanks{University of Central Florida}\\
{\tt\small xu979022@ucf.edu}
\and
Alejandro Aparcedo\footnotemark[1] \\
{\tt\small aaparcedo@ucf.edu}
\and
Young Kyun Jang\\
{\tt\small kyun0914@gmail.com}
\and
Ser-Nam Lim\footnotemark[1]\\
{\tt\small sernam@ucf.edu}
}

\begin{document}
\maketitle
\begin{abstract}
Recent advances in instruction tuning have led to the development of State-of-the-Art Large Multimodal Models (LMMs). Given the novelty of these models, the impact of visual adversarial attacks on LMMs has not been thoroughly examined. We conduct a comprehensive study of the robustness of various LMMs against different adversarial attacks, evaluated across tasks including image classification, image captioning, and Visual Question Answer (VQA). We find that in general LMMs are not robust to visual adversarial inputs. However, our findings suggest that context provided to the model via prompts—such as questions in a QA pair—
helps to mitigate the effects of visual adversarial inputs. Notably, the LMMs evaluated demonstrated remarkable resilience to such attacks on the ScienceQA task with only an 8.10\% drop in performance compared to their visual counterparts which dropped 99.73\%. We also propose a new approach to real-world image classification which we term query decomposition. By incorporating existence queries into our input prompt we observe diminished attack effectiveness and improvements in image classification accuracy. This research highlights a previously under explored facet of LMM robustness and sets the stage for future work aimed at strengthening the resilience of multimodal systems in adversarial environments.
\end{abstract}

\section{Introduction}
\label{sec:intro}

Large Multi-modal Models (LMMs) have demonstrated remarkable abilities in a range of applications, from image classification and Visual Question Answering (VQA) to image captioning and semantic segmentation~\cite{alayrac2022flamingo, lai2023lisa, liu2023visual, dai2023instructblip, li2023blip2}. These models excel in generalizing to new domains with data-efficient solution, a feat attributed to advancements in Instruction Tuning \cite{wei2022finetuned}. Such techniques, traditionally applied to text-only models, have now been extended to multi-modal models, opening new avenues for efficient fine-tuning with significantly less data \cite{liu2023visual, dai2023instructblip}. 

 \begin{figure}[ht]
    \centering
    \includegraphics[width=\columnwidth]{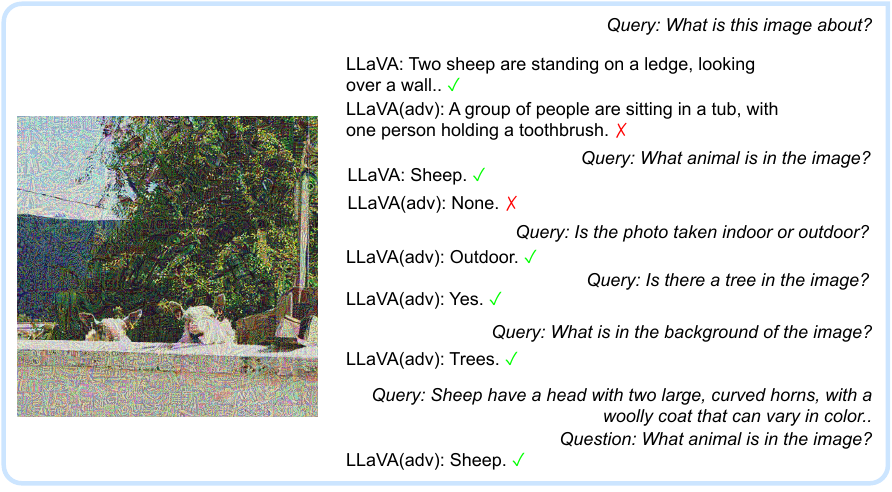}
    \caption{QA pairs for LLaVA~\cite{liu2023visual} given an adversarial image. ``LLaVA" and ``LLaVA(adv)" refer to LLaVA's response to the user query with clean and adversarial image, respectively. For the readers, there are two sheep in the scene, and the adversarial attack was based on maximizing the distance between the image and the text ``a photo of a \textit{sheep}". In the first two QA pairs, we can see that LLaVA(adv)'s answer is completely wrong. However, it can still answer the following questions correctly, because they are not pertinent to the object being attacked (sheep). Also note the contrast between the second and last QA pairs. LLaVA(adv) answers the question correctly after additional context has been provided. These observations help drove some of the findings in this paper. Source: COCO~\cite{lin2014microsoftcoco}}
    \label{fig:llava_response}
\end{figure}

Despite the recent advancements in LMMs, the impact of adversarial examples still remains under explored. Typically adversarial examples are generated end-to-end, targeting the final loss of the whole model, and focusing on a single modality. However, in the era of combining different pre-trained models with additional projectors or adaptors \cite{liu2023visual, zhu2023minigpt4, chen2023shikra}, it is imperative to reevaluate the effectiveness of these adversarial approaches. For example, let's consider LLaVA \cite{liu2023visual} which uses CLIP as its visual component and LLAMA as text component (with some additional projector to bridge the gap), will an attack on one of the two components compromise its overall performance?

From a practical perspective, given the substantial size of LMMs, attacking the entire model is often prohibitively expensive \cite{carlini2023aligned}, making the above question an increasingly important one to answer since traditional adversarial attacks are better developed and computationally cheaper.
Specifically, in this paper, we question the efficacy of adversarial attacks against visual encoders when they serve as input to subsequent LLMs. This gap in understanding raises critical questions about the susceptibility of LMMs to adversarial attacks, especially when only the visual encoder is targeted. Given their sophisticated dual-model composition, the question arises: \textit{Can an attack on the visual encoder effectively compromise the entire LMM?}


 Recent works~\cite{qi2023visual, carlini2023aligned} on visual adversarial attacks against LMMs typically focus on the safety and alignment aspects of the model. For example, Qi et al~\cite{qi2023visual} and Carlini et al~\cite{carlini2023aligned} both show that it is possible to generate a visual adversarial example that ``jailbreak" the LMM. Nonetheless, a systematic study on the impact of visual adversaries on LMMs is still missing.

We conduct a comprehensive analysis on the robustness of current LMMs under various adversarial attacks, tasks and datasets. Our investigation reveals that LMMs are not robust to adversarial visual perturbations in contexts where no additional textual information is provided, such as in COCO\cite{lin2014microsoftcoco} classification (without context) or COCO captioning tasks. Conversely, the presence of context seems to bolster LMM robustness, as seen in tasks like COCO classification (with context). In cases where the attack does not directly target the core aspects of the task, such as in VQA, LMMs display a degree of inherent robustness.
%
%
This paper reveals the following findings:
\begin{itemize}
    \item LMMs are generally vulnerable to adversarial visual perturbations, even if such perturbations are generated only w.r.t. the visual model. 
    \item LMMs shows decent robustness in VQA tasks. We find that visual attacks are less effective when the VQA question query involves different visual contents from what is being attacked.
    \item Adding additional textual context notably improves LMMs' robustness against visual adversarial input.
    \item Based on the above findings, we devise a context-augmented image classification scheme that shows non-trivial increase in robustness.
\end{itemize}


\section{Related Work}
\label{sec:related}

\textbf{Large Multimodal Models (LMMs).} 
%
Large Multimodal Models (LMMs)\cite{liu2023visual, li2023blip2, chen2023shikra, zhu2023minigpt4, qwen} typically comprise a visual model, a pre-trained Large Language Model (LLM), and a projector model designed to bridge the modality gap between images and text. Prominent among these models are LLaVA\cite{liu2023visual} and InstructBLIP~\cite{dai2023instructblip}, which represent the current state-of-the-art in LMMs. LLaVA integrates the CLIP visual encoder with the Vicuna LLM~\cite{vicuna2023}, employing a simple linear projector subsequent to the visual model for transforming visual representations into the language embedding space. Conversely, BLIP2-based models~\cite{li2023blip2, dai2023instructblip, zhu2023minigpt4} utilize the EVA-CLIP visual encoder, alongside a Q-former equipped with learnable query vectors to bridge the visual and textual modalities. Both LLaVA and BLIP2-based models, among others, have demonstrated remarkable capabilities in a variety of vision-language tasks, underscoring their versatility and effectiveness.

\noindent\textbf{Adversarial attacks.} 
Adversarial attacks are designed to subtly manipulate inputs in a way that is typically imperceptible to humans, yet can lead neural networks to produce erroneous outputs~\cite{Szegedyintrig, Biggio_2013, carlini2017, madry2018towards, pmlr-v119-croce20b, obfuscated-gradients}. These attacks are broadly classified into two categories: white-box attacks~\cite{Szegedyintrig, carlini2017, obfuscated-gradients}, where the adversary has complete access to the model parameters, and black-box attacks~\cite{Papernot_blackbox,Su2017OnePA}, where the adversary possesses only limited information such as output logits or labels.

While the primary focus of adversarial attack research has historically been on image classification, recent studies have demonstrated the feasibility of constructing adversarial examples in textual domains. These examples can be generated either heuristically~\cite{jia-liang-2017-adversarial, alzantot-etal-2018-generating, li2021adversarialvqa} or through discrete optimization techniques~\cite{ebrahimi-etal-2018-hotflip, wallace-etal-2019-universal}.

\noindent\textbf{LMMs and Adversarial Examples.} 
While extensive research has been conducted on adversarial attacks in both visual and textual domains, the impact of these attacks on current LMMs remains relatively unexplored. Recent studies ~\cite{qi2023visual, carlini2023aligned, zou2023universal, wei2023jailbroken, liu2023jailbreaking, rao2023tricking} demonstrate the feasibility of creating adversarial examples that effectively "jailbreak" LMMs from both visual~\cite{qi2023visual, carlini2023aligned} and textual~\cite{zou2023universal, wei2023jailbroken, liu2023jailbreaking, rao2023tricking} inputs, using either gradient-based approaches~\cite{qi2023visual, carlini2023aligned} or prompt engineering~\cite{wei2023jailbroken, liu2023jailbreaking, rao2023tricking}. These examples are capable of inducing LMMs to produce harmful content, thereby bypassing the safety measures implemented during model alignment, such as instruction tuning or Reinforcement Learning from Human Feedback (RLHF). However, while these studies predominantly address the safety concerns, potential harmfulness, and the associated dangers of LMMs, our research shifts the focus towards systematically examining the accuracy of LMMs in performing various tasks under the influence of visual adversarial attacks.

\section{Method}

\subsection{Problem Statement}
In this study, we focus on gradient-based white-box adversarial attacks~\cite{carlini2017, madry2018towards,pmlr-v119-croce20b}. These methods hinge on the computation of the gradient to ascertain the most effective direction in which to modify the input so as to deceive the model, while satisfying the $L_p$ constraint. Formally, given input-label pair $x,y$ and the model denoted by $f$, we want to find the adversarial perturbation $\delta$ s.t. $f(x+\delta) \neq y$ confined to some $L_p$ bounds. For PGD, we maximize $L(f(x+\delta), y)$ while satisfying $||\delta||_{\infty} < \epsilon$, where $\epsilon$ is the radius of the $L_{\infty}$ ball, and $L$ is the Cross-Entropy loss in our case. For CW, we maximize $||\delta||_p + c \cdot g(x+\delta)$, subject to $x + \delta \in [0, 1]^n$, where $g(x+\delta)=\max(f(x+\delta)_y - \max\{f(x+\delta)_i : i \neq y\}, -\kappa)$, and $\kappa$ is the confidence parameter. 

\subsection{Attacks}
 We choose PGD and CW as two representatives of strong gradient-based attacks, along with APGD as a variant of PGD. Additionally, we experiment with two parameter settings of each attack: normal and strong, based on perceptibility of the perturbations. Under the normal setting, we set the constraint for CW to 20, and epsilon for PGD/APGD to 8/255, as used in prior works~\cite{obfuscated-gradients,xie2017mitigating}. Under the strong setting, we set epsilon for PGD and APGD to 0.2, and constraint to 100 for CW. All the attacks are generated solely w.r.t. the image encoder, leaving the LLM untouched. Detailed parameters can be found in Table~\ref{attack_params}.
 
\begin{figure}[ht]
    \centering
    \includegraphics[width=0.7\columnwidth]{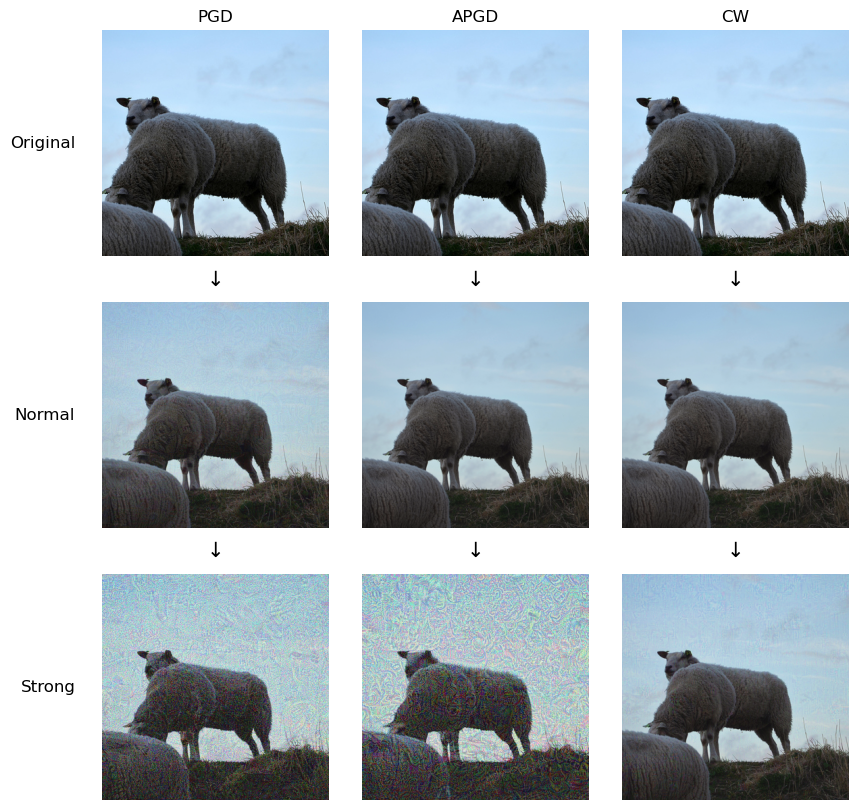}
    \caption{A sample CLIP's adversarial image, generated by PGD, APGD and CW, under Normal and Strong attack parameter settings. Image source: COCO 2014val. Note that under strong attack, the adversarial perturbations become very obvious under PGD and APGD, and are expected to cause a higher degree of performance degradation.}
    \label{fig:adv_sample}
\end{figure}

 Figure~\ref{fig:adv_sample} shows a sample adversary generated using different attack methods and under different degree of attack strength. In the normal setting, the adversarial perturbation is almost imperceptible, but become obvious under strong setting for PGD and APGD. Perturbations generated by CW remains imperceptible even under the strong setting. For brevity, in the follow sections, we use N and S to represent normal and strong setting, respectively. For example, APGD-S stands for APGD attack under strong setting. 

\begin{table}
\small
\centering
\begin{tabular}{lcccccc}
 \toprule
Method & steps & step size & $\epsilon$ & Dist. & $c$ & $\kappa$ \\
 \midrule
 \multicolumn{7}{c}{Normal} \\
PGD & 20 & 2/255 & 8/255 & $L_{\infty}$ & - & - \\
APGD & 20 & - & 8/255 & $L_{\infty}$ & - & -\\
CW & 50 & 0.01 & - & $L_{2}$ & 20 & 0 \\
\midrule
\multicolumn{7}{c}{Strong}\\
PGD & 40 & 2/255 & 0.2 & $L_{\infty}$ & - & - \\
APGD & 40 & - & 0.2 & $L_{\infty}$ & - & -\\
CW & 75 & 0.05 & - & $L_{2}$ & 100  & 0 \\
 \bottomrule
\end{tabular}
\caption{Parameters for the attacks under Normal (N) and Strong (S) settings. Dist. refers to distance measure, $c$ and $\kappa$ refers to the constraint and confidence parameter in ~\cite{carlini2017}.}
\label{attack_params}
\end{table}

\subsection{Models}
In our study, we selected three state-of-the-art LMM models for evaluation: LLaVA1.5\cite{liu2023improved} integrated with the Vicuna13B language model, BLIP2 combined with the Flan T5 XXL\cite{chung2022scaling} language model, and InstructBLIP~\cite{dai2023instructblip}, also utilizing Vicuna13b. These models exhibit distinct characteristics in their configurations. LLaVA1.5 and InstructBLIP both employ the Vicuna13B language model; however, they differ in their image encoders and methodologies for merging image and text encodings, with LLaVA1.5 directly inserting the projected visual tokens into text tokens, versus InstructBLIP's Q-former architecture. BLIP2 and InstructBLIP share similar image encoders and the Q-former architecture but diverge in their language model choices and training protocols: BLIP2 employs the Encoder-Decoder based Flan T5 XXL, while InstructBLIP uses the Decoder-only Vicuna13B. We believe that such a selection of models allows for a diverse yet controlled set of experimental conditions. 
In the rest of the paper, we use LLaVA, BLIP2-T5 and InstructBLIP to refer to LLaVA1.5 Vicuna13B, BLIP2 Flan T5XXL, and InstructBLIP Vicuna13B, respectively. 

\subsection{Tasks \& Adversarial generation}
We consider three popular visual tasks for evaluating visual adversarial impact on LMMs: image classification, caption retrieval and VQA. Since we are interested in LMMs' robustness against visual adversaries, we generate adversarial samples w.r.t. the image encoder of the LMM: CLIP image encoder for LLaVA, EVA-CLIP image encoder for BLIP2 and InstructBLIP. We use CLIP text encoder and the text encoder from BLIP's Q-former to compute the text embeddings for their corresponding image encoder. For PGD and APGD, we maximize the Cross-Entropy loss between the model logits and the ground-truth label. For CW, we minimize the sum of the $l_2$ distance of the perturbation $\delta$ and the $f$-function from the original paper~\cite{carlini2017}. Detailed procedures for task-specific adversarial generation are given below.




 \begin{figure}[ht]
    \centering
    \includegraphics[width=\columnwidth]{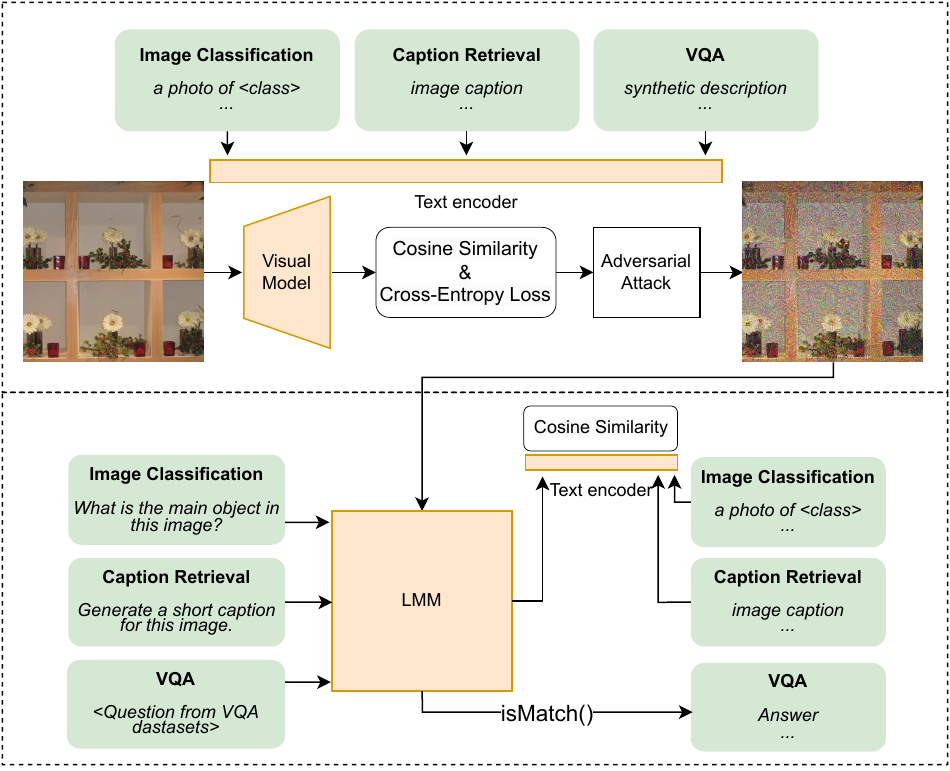}
    \caption{Overview of our procedure for attack generation and evaluation over image classification, caption retrieval, and VGA. Top: overview of attack generation for the three tasks; bottom: evaluation procedure for LMM on the three tasks.}
    \label{fig:diagram}
\end{figure}

\subsubsection{Image Classification}
We use COCO \cite{lin2014microsoftcoco} 2014 validation split (2014val), with class annotations from~\cite{DHD}, to evaluate robustness on classification. We first use the text encoder to encode the text class labels in the format of ``a photo of \textless class\textgreater". Then, we compute the class-wise cosine-similarity between the image encodings and encoded class labels and use the result as the class logits for adversarial generation and evaluation. To evaluate LMMs on classification, we first prompt LMMs to generate a one-word response of the main object in the image. For LLaVA we use the prompt: ``What is the main object in this image?\textbackslash nAnswer in a single word or phrase." For BLIP2-T5 and InstructBLIP we use ``Question: what is the main object in this image? Short answer: ". We then format the answer with ``a photo of \textless answer\textgreater", encode with the text encoder and compute cosine similarity against the class label encodings to perform classification.

\subsubsection{Caption retrieval}
\label{subsec:caption-retrieval}
We use COCO captioning dataset~\cite{coco_caption} 2014val for evaluating caption retrieval robustness. To generate visual adversarial samples for caption retrieval, we first use the text encoder to encode 5 captions per image, and then use their mean as the text encodings for each image. Then, we compute cosine similarity between image and text encodings and use the result as the image-wise logits for adversarial generation. To evaluate the caption retrieval for CLIP and EVA-CLIP encoders, we compute cosine similarity between the image encodings and all captions' text encodings to perform retrieval. To evaluate caption retrieval for LMMs, we first prompt LMMs to generate a caption for the image. For LLaVA we use the prompt: ``Describe this image in a short sentence." For BLIP2-T5 and InstructBLIP we use the prompt: ``Question: what is this image about? Short answer: " Then we encode the generated caption and compute cosine similarity against all captions' text encodings to perform retrieval.

\subsubsection{VQA}
We evaluate LMM robustness on five popular VQA datasets: VQA V2~\cite{goyal2017making}, ScienceQA-Image~\cite{lu2022learn}, TextVQA~\cite{singh2019towards}, POPE~\cite{li2023evaluating} and MME~\cite{fu2023mme}. For VQA V2, we follow the same adversarial generation procedure as in classification task. For all other datasets, since no ground-truth label is present, we first prompt LLaVA with ``What is this image about?\textbackslash nAnswer in one sentence." to generate synthetic caption for each image, and follow the same procedure in caption retrieval task for generating adversaries. We follow the official evaluation procedures for each VQA datasets.

\section{Experimental Results and Analysis}

We show our experimental results and analysis in the following sections. We report both LMMs' accuracy as well as the image encoder's accuracy on the task that was used to generate adversaries.
We adopt the notations Pre, $\text{Post}_\text{N}$ and $\text{Post}_\text{S}$ to refer to accuracy for pre-attack, post-attack under normal setting, and post-attack under strong setting, respectively.

\begin{table}
\small
\centering
\begin{tabular}{lccccc}
 \toprule
Model & Attack & Pre & $\text{Post}_\text{N}$ & $\text{Post}_\text{S}$ \\
 \midrule
  \multicolumn{5}{c}{Visual Encoder Acc @1 (\%)} \\
CLIP & PGD & 63.32 & 11.78\decr{81} & 0.48\decr{99} \\
CLIP & APGD & 63.32 & 4.2\decr{93} & 0.02\decr{100} \\
CLIP & CW & 63.32 & 13.86 \decr{78} & 0.94\decr{99} \\
EVA-CLIP & PGD & 73.18 & 1.02\decr{99} & 0.0\decr{100} \\
EVA-CLIP & APGD & 73.18 & 0.46\decr{99} & 0.0\decr{100} \\
EVA-CLIP & CW & 73.18 & 19.43 \decr{73} & 3.74\decr{95} \\
\midrule
 \multicolumn{5}{c}{Image-to-Text Recall @1 (\%)} \\
CLIP & PGD & 57.72 & 10.4\decr{82} & 0.4\decr{99} \\
CLIP & APGD & 57.72 & 12.92\decr{78} & 7.44\decr{87} \\
CLIP & CW & 57.72 & 34.94\decr{39} & 24.94\decr{57} \\
EVA-CLIP & PGD & 64.06 & 1.06\decr{98} & 0.06\decr{100} \\
EVA-CLIP & APGD & 64.06 & 9.14\decr{86} & 8.32\decr{87} \\
EVA-CLIP & CW & 64.06 & 42.06\decr{34} & 31.72\decr{50} \\
\midrule
 \multicolumn{5}{c}{LLM Answer-to-Text Recall @1 (\%)} \\
LLaVA & PGD & 36.58 & 13.1\decr{64} & 3.76\decr{90} \\
LLaVA & APGD & 36.58 & 15.7\decr{57} & 7.88\decr{78} \\
LLaVA & CW & 36.58 & 32.96\decr{10} & 29.84\decr{18} \\
BLIP2-T5 & PGD & 32.34 & 1.4\decr{96} & 0.1\decr{100} \\
BLIP2-T5 & APGD & 32.34 & 4.52\decr{86} & 3.62\decr{89} \\
BLIP2-T5 & CW & 32.34 & 23.12\decr{29} & 17.02\decr{47} \\
InstructBLIP & PGD & 37.82 & 1.9\decr{95} & 0.18\decr{100} \\
InstructBLIP & APGD & 37.82 & 5.56\decr{85} & 4.3\decr{89} \\
InstructBLIP & CW & 37.82 & 27.44\decr{27} & 20.74\decr{45} \\
 \bottomrule
\end{tabular}
\caption{Top-1 caption retrieval result for COCO caption 2014 validation dataset. Refer to Sec.~\ref{subsec:caption-retrieval}. ``Visual Encoder Accuracy" refers to CLIP/EVA-CLIP accuracy on successfully retrieving captions that are closed to the mean caption encoding given the image encoding. ``Image-to-Text Recall @1" is recall@1 of retrieving correctly one of the five captions for the given image. LLM Answer-to-Text recall is the same except the query is the LMMs' answers. Numbers in parenthesis show \% change w.r.t. the Pre-attack accuracy.}
\label{coco_caption}
\end{table}

\subsection{Are LMMs Robust Against Adversarial Visual Input?}
\label{subsec:caption-attack}
To investigate the impact of adversarial visual inputs on LMMs, our initial analysis focuses on the caption retrieval task. This task serves as a measure of the LMMs' overall comprehension of visual inputs. The results of this analysis, conducted on COCO 2014val, are presented in Table~\ref{coco_caption} -- please refer to the caption to comprehend the significance of each metric. Under the third section, the data distinctly illustrates a significant decrease in post-attack accuracy across all three LMMs when subjected to both PGD and APGD attacks, under both normal and strong settings. For instance, 
under PGD-N, the Top-1 recall rate for InstructBLIP declines to 1.9\%, and further diminishes to 0.18\% under PGD-S. Since these accuracies are roughly on the same level as the CLIP/EVA-CLIP accuracy post attack, as shown under the second section of the table, the additional LLM appended to the image encoder did not bring notable robustness, indicating that LMMs lack robustness against visual adversarial perturbations. In other words, the visual perturbations are capable of substantially undermining the LMMs' effectiveness, even though they are not generated through an end-to-end process on the LMMs' text generation loss.


\begin{table*}
\centering
\begin{tabular}{lcccccccc}
 \toprule
 & & & \multicolumn{3}{c}{VQA Acc (\%)} & \multicolumn{3}{c}{Visual Encoder Acc (\%)} \\
Model & Dataset & Attack & Pre & $\text{Post}_\text{N}$ & $\text{Post}_\text{S}$ & Pre & $\text{Post}_\text{N}$ & $\text{Post}_\text{S}$ \\
 \cmidrule(lr){1-3} \cmidrule(lr){4-6} \cmidrule(lr){7-9}
LLaVA & ScienceQA(image) & PGD & 71.59 & 68.77 \decr{3} & 64.75 \decr{9} & 42.92 & 10.08 \decr{77} & 0.92 \decr{98} \\
LLaVA & ScienceQA(image) & APGD & 71.59 & 69.81 \decr{2} & 68.22 \decr{4} & 42.83 & 5.68 \decr{87} & 0.06 \decr{99} \\
LLaVA & ScienceQA(image) & CW & 71.59 & 71.69 \incr{0.1} & 71.34 \decr{0.3} & 42.95 & 12.76 \decr{70} & 0.03 \decr{99} \\
BLIP2-T5 & ScienceQA(image) & PGD & 74.71 & 69.71 \decr{6} & 63.06 \decr{15} & 46.40 & 1.05 \decr{98} & 0.00 \decr{100} \\
BLIP2-T5 & ScienceQA(image) & APGD & 74.71 & 73.62 \decr{1} & 72.88 \decr{2} & 46.40 & 1.02 \decr{98} & 0.00 \decr{100} \\
BLIP2-T5 & ScienceQA(image) & CW & 74.71 & 74.71 \decr{0} & 74.37 \decr{0.4} & 46.40 & 12.92 \decr{72} & 0.03 \decr{99} \\
InstructBLIP & ScienceQA(image) & PGD & 45.08 & 40.66 \decr{10} & 39.67 \decr{12} & 46.40 & 1.05 \decr{98} & 0.00 \decr{100} \\
InstructBLIP & ScienceQA(image) & APGD & 45.08 & 42.75 \decr{5} & 42.34 \decr{3} & 46.40 & 1.02 \decr{98} & 0.00 \decr{100} \\
InstructBLIP & ScienceQA(image) & CW & 45.08 & 44.88 \decr{0.4} & 43.93 \decr{0.3} & 46.40 & 12.92 \decr{72} & 0.03 \decr{99} \\
 \cmidrule(lr){1-3} \cmidrule(lr){4-6} \cmidrule(lr){7-9}
LLaVA & VQA V2 & PGD & 78.43 & 64.38 \decr{18} & 51.22 \decr{35} & 89.21 & 31.00 \decr{65} & 6.56 \decr{93} \\
LLaVA & VQA V2 & APGD & 78.43 & 67.41 \decr{14} & 44.60 \decr{43} & 89.21 & 30.01 \decr{66} & 0.15 \decr{99} \\
LLaVA & VQA V2 & CW & 78.43 & 76.79 \decr{2.1} & 75.16 \decr{4.1} & 89.21 & 44.92 \decr{50} & 0.04 \decr{99} \\
BLIP2-T5 & VQA V2 & PGD & 66.94 & 50.60 \decr{24} & 42.43 \decr{37} & 94.13 & 19.12 \decr{80} & 0.23 \decr{99} \\
BLIP2-T5 & VQA V2 & APGD & 66.94 & 52.51 \decr{22} & 40.69 \decr{39} & 94.13 & 14.71 \decr{84} & 0.01 \decr{99} \\
BLIP2-T5 & VQA V2 & CW & 66.94 & 63.69 \decr{4.8} & 58.75 \decr{12} & 94.12 & 51.60 \decr{45} & 0.06 \decr{99} \\
InstructBLIP & VQA V2 & PGD & 76.07 & 56.33 \decr{26} & 42.77 \decr{44} & 94.13 & 19.12 \decr{80} & 0.23 \decr{99} \\
InstructBLIP & VQA V2 & APGD & 76.07 & 58.83 \decr{22} & 39.60 \decr{48} & 94.13 & 14.71 \decr{84} & 0.01 \decr{99} \\
InstructBLIP & VQA V2 & CW & 76.07 & 73.02 \decr{4.0} & 66.10 \decr{13} & 94.12 & 51.60 \decr{45} & 0.06 \decr{99} \\
 \cmidrule(lr){1-3} \cmidrule(lr){4-6} \cmidrule(lr){7-9}
LLaVA & TextVQA & PGD & 62.14 & 51.44 \decr{17} & 40.27 \decr{35} & 69.32  & 10.30 \decr{85} & 0.41 \decr{99} \\
LLaVA & TextVQA & APGD & 62.14 & 54.23 \decr{12} & 42.88 \decr{31} & 69.32 & 6.73 \decr{90} & 0.00 \decr{100} \\
LLaVA & TextVQA & CW & 62.14 & 60.88 \decr{2} & 59.71 \decr{4} & 69.38 & 18.00 \decr{74} & 0.03 \decr{99} \\
BLIP2-T5 & TextVQA & PGD & 45.14 & 38.46 \decr{14} & 29.82 \decr{34} & 68.97 & 0.44 \decr{99} & 0.00 \decr{100} \\
BLIP2-T5 & TextVQA & APGD & 45.14 & 39.94 \decr{11} & 32.41 \decr{28} & 68.97 & 0.63 \decr{99} & 0.00 \decr{100} \\
BLIP2-T5 & TextVQA & CW & 45.14 & 44.28 \decr{2} & 38.58 \decr{14} & 69.01 & 12.51 \decr{82} & 0.03 \decr{99} \\
InstructBLIP & TextVQA & PGD & 35.23 & 26.99 \decr{23} & 18.11 \decr{48} & 68.97 & 0.44 \decr{99} & 0.00 \decr{100} \\
InstructBLIP & TextVQA & APGD & 35.23 & 28.00 \decr{20} & 20.10 \decr{43} & 68.97 & 0.63 \decr{99} & 0.00 \decr{100} \\
InstructBLIP & TextVQA & CW & 35.23 & 33.95 \decr{3} & 25.46 \decr{27} & 69.01 & 12.51 \decr{82} & 0.03 \decr{99} \\
 \cmidrule(lr){1-3} \cmidrule(lr){4-6} \cmidrule(lr){7-9}
LLaVA & POPE & PGD & 85.55 & 73.13 \decr{14} & 58.97 \decr{31} & 80.00 & 7.20 \decr{91} & 0.2 \decr{99} \\
LLaVA & POPE & APGD & 85.55 & 73.80 \decr{13} & 65.00 \decr{24} & 80.00 & 4.40 \decr{94} & 0.00 \decr{100} \\
LLaVA & POPE & CW & 85.55 & 83.07 \decr{2} & 83.27 \decr{2} & 80.00 & 21.40 \decr{73} & 0.80 \decr{99} \\
BLIP2-T5 & POPE & PGD & 77.10 & 62.67 \decr{18} & 55.50 \decr{28} & 87.40 & 0.00 \decr{100} & 0.00 \decr{100} \\
BLIP2-T5 & POPE & APGD & 77.10 & 65.20 \decr{15} & 55.30 \decr{28} & 87.40 & 0.20 \decr{99} & 0.00 \decr{100} \\
BLIP2-T5 & POPE & CW & 77.10 & 75.87 \decr{1} & 75.20 \decr{2} & 87.40 & 15.80 \decr{81} & 3.80 \decr{95} \\
InstructBLIP & POPE & PGD & 82.83 & 64.57 \decr{22} & 52.20 \decr{37} & 87.40 & 0.00 \decr{100} & 0.00 \decr{100} \\
InstructBLIP & POPE & APGD & 82.83 & 66.93 \decr{19} & 52.00 \decr{37} & 87.40 & 0.20 \decr{99} & 0.00 \decr{100} \\
InstructBLIP & POPE & CW & 82.83 & 80.93 \decr{2} & 80.27 \decr{3} & 87.40 & 15.80 \decr{82} & 3.80 \decr{95} \\
 \cmidrule(lr){1-3} \cmidrule(lr){4-6} \cmidrule(lr){7-9}
LLaVA & MME & PGD & 1,536 & 1,187 \decr{22} & 927 \decr{39} & 65.79 & 12.25 \decr{81} & 0.91 \decr{98} \\
LLaVA & MME & APGD & 1,536 & 1,283 \decr{16} & 818 \decr{46} & 65.79 & 7.29 \decr{88} & 0.10 \decr{99} \\
LLaVA & MME & CW & 1,536 & 1,521 \decr{1} & 1491 \decr{3} & 65.79 & 17.51 \decr{73} & 3.04 \decr{95} \\
BLIP2-T5 & MME & PGD & 1,114 & 759 \decr{32} & 591 \decr{47} & 73.38 & 2.43 \decr{96} & 0.00 \decr{100} \\
BLIP2-T5 & MME & APGD & 1,114 & 777 \decr{30} & 628 \decr{43} & 73.38 & 2.02 \decr{97} & 0.00 \decr{100} \\
BLIP2-T5 & MME & CW & 1,114 & 1058 \decr{5} & 1026 \decr{8} & 73.38 & 18.62 \decr{74} & 6.17 \decr{91} \\
InstructBLIP & MME & PGD & 1,248 & 704 \decr{30} & 703 \decr{43} & 73.38 & 2.43 \decr{96} & 0.00 \decr{100} \\
InstructBLIP & MME & APGD & 1,248 & 1,002 \decr{19} & 751 \decr{40} & 73.38 & 2.02 \decr{97} & 0.00 \decr{100} \\
InstructBLIP & MME & CW & 1,248 & 1,205 \decr{3} & 1,170 \decr{6} & 73.38 & 18.62 \decr{74} & 6.17 \decr{91} \\
 \bottomrule
\end{tabular}
\caption{Results on VQA datasets. We attack CLIP and EVA-CLIP visual encoders to generate adversarial examples for LLaVA and BLIP2-T5/InstructBLIP, respectively. Adversarial examples are used as input image along with question as input text. ``VQA Accuracy" refers to the performance of each LMM; ``Visual Encoder Accuracy" refers to the accuracy of the visual encoder on image-to-text retrieval, which is used for generating visual adversaries for VQA. Numbers in parenthesis show \% change w.r.t. the Pre-attack accuracy.} 
\label{vqa}
\end{table*}

\begin{figure}[ht]
    \centering
    \includegraphics[width=\columnwidth]{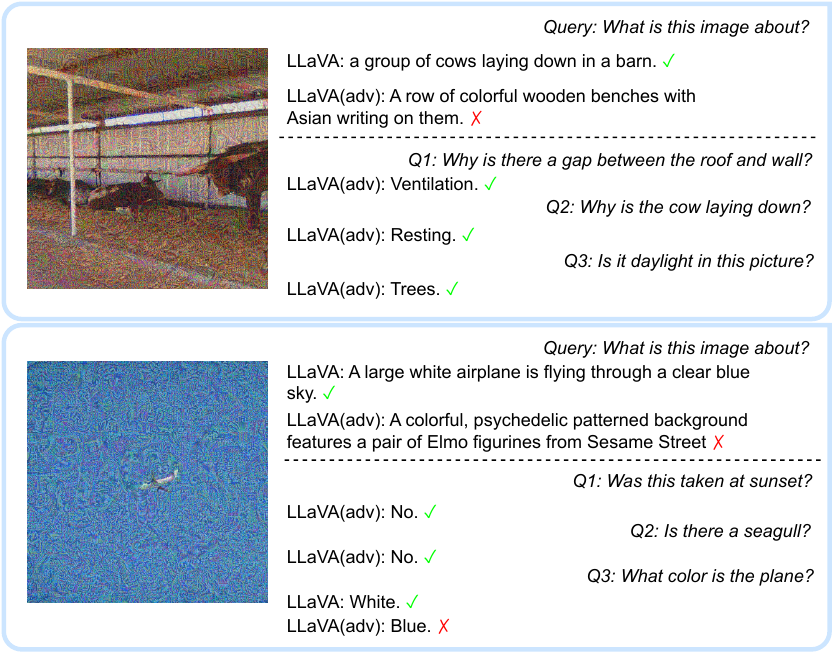}
    \caption{Two sample adversarial images from COCO 2014val, generated under APGD $\text{Post}_\text{S}$. ``LLaVA" and ``LLaVA(adv)" refer to LLaVA's responses using the clean Pre-attack and post-attack image, respectively. Above the dotted line in each cell, we query LLaVA for the general description; below the dotted line are questions taken from VQA V2 dataset.}
    \label{fig:llava_response}
\end{figure}

\subsection{Evaluating LMMs' VQA Performance}
In this section, we detail the experimental outcomes of the LMMs in VQA tasks under adversarial visual attacks. The primary results are summarized in Table~\ref{vqa}. Our results indicate a noteworthy deviation from what we have observed about the caption retrieval task in Sec.~\ref{subsec:caption-attack}, which did not show that LMMs possess any robustness against visual adversaries. Based on the results from Table~\ref{vqa}, all three LMMs being evaluated exhibit considerable resilience in various VQA datasets, despite the significant decrease in adversarial accuracy of their corresponding visual encoders, as shown under the ``Visual Encoder Accuracy" columns. For instance, with the ScienceQA dataset, the $\text{Post}_\text{N}$ ``Visual Encoder Accuracy" plummeted below 1\% for all three types of attacks, and for both the CLIP and BLIP visual encoders. However, the accuracy of all three LMMs decreased by less than 7\% compared to their pre-attack accuracy. 

What could be the cause of such discrepancies in LMMs' robustness between the VQA and caption retrieval tasks? We make two conjectures: 
\begin{enumerate}
    \item The robustness of LMMs depends on whether the query is about what is being attacked. Since the attack target for generating visual adversarial samples is what is being described in the image description, then intuitively those aspects not mentioned in the description shall be less affected by the attack.
    \item Additional contexts (e.g., contexts in ScienceQA's questions) aid in LMMs' robustness.
\end{enumerate}
We will experimentally support the two claims in the following sections.

\subsection{Visual Adversarial Attacks are not Universal to LMMs}
In this section, we present an empirical analysis demonstrating that while LMMs are not inherently resilient to visual adversarial attacks, as evidenced by their performance in caption retrieval tasks, they are capable of delivering correct responses when the query's focus differs from the target of the attack. To illustrate this, we take the Visual Question Answering (VQA) V2 dataset as a case study. Here, we generate adversarial images using the text label ``a photo of \textless class\textgreater", with the attack primarily aimed at the central object of the image. We observe that the adversarial attack's effectiveness is heightened when the query, during evaluation, pertains to the same target – the principal object in the image. Conversely, the attack's impact diminishes when the query relates to different aspects of the image.

\begin{figure}[ht]
    \centering
    \includegraphics[width=\columnwidth]{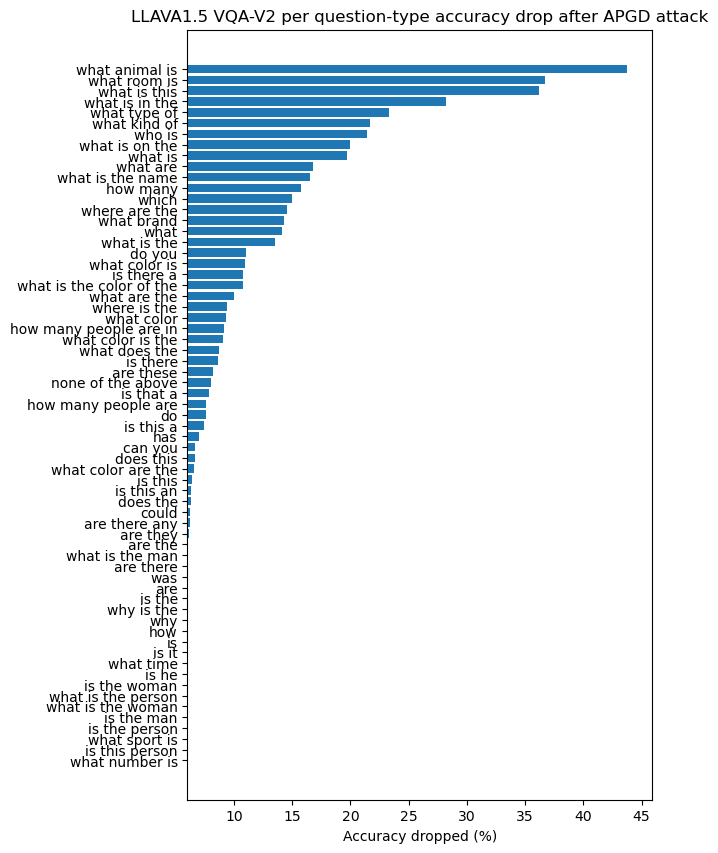}
    \caption{LLaVA's VQA accuracy drop after APGD attack under the normal attack setting. Y-axis represents question types, and X-axis represents accuracy dropped (\%).}
    \label{fig:llava_norm_vqav2_drop}
\end{figure}

In Figure~\ref{fig:llava_response}, we show LLaVA's responses to queries on two adversarial images under APGD-S. When querying about the general description of the image, it is clear that LLaVA's post-attack answers are completely deviated from what the image is about; however, below the dotted line, LLaVA can still answer most questions correctly. We conjecture such phenomenon is either because LLaVA can ``guess" the answer directly from the context (e.g., Q2-top ``Why is the cow laying down?" -- ``Resting"). This is coherent with LMMs' high ``robustness" on the ScienceQA dataset, in which the texts themselves are often sufficient to find the answer. On the other hand, it is because these questions are not directly querying the object or its attributes, but rather the peripheral aspects of the image (e.g., Q1-bottom ``Was this taken at sunset?"). The only incorrectly-answered question is Q3-bottom ``What color is the plane?". LLaVA answers it incorrectly as the query is asking about the object attribute (color), which has been corrupted by the attack.

In Figure~\ref{fig:llava_norm_vqav2_drop}, we plot LLaVA's per-question type accuracy drop under APGD-N. We can clearly see that accuracy drops the most on questions asking `What' -- what room/animal/color -- about the object and its direct attributes. The accuracy drop quickly diminishes for question asking `Is/Has/Can' etc. These questions are typically querying the peripheral aspects of the image instead of the main object, 
and actually require more complex reasoning and understanding to answer correctly, yet they boast much lower accuracy drop. This result reaffirms our conjecture that LMMs are robust when the question is not querying what is being attacked.

\begin{table}
\small
\centering
\begin{tabular}{lcccc}
 \toprule
Model & Attack & Pre@1 & $\text{Post}_\text{N}$@1 & $\text{Post}_\text{S}$@1 \\
 \midrule
 \multicolumn{5}{c}{Visual Encoder Acc (\%)} \\
CLIP & PGD & 89.21 & 31.0\decr{65} & 6.56\decr{93} \\
CLIP & APGD & 89.21 & 30.01\decr{66} & 0.15\decr{100} \\
CLIP & CW & 89.21 & 44.92\decr{50} & 0.04\decr{100} \\
EVA-CLIP & PGD & 94.13 & 19.12\decr{80} & 0.23\decr{100} \\
EVA-CLIP & APGD & 94.13 & 14.71\decr{84} & 0.01\decr{100} \\
EVA-CLIP & CW & 94.12 & 51.6\decr{45} & 0.06\decr{100} \\
\midrule
\multicolumn{5}{c}{LMM Acc (\%)}  \\
LLaVA  & PGD & 87.51 & 48.25\decr{45} & 22.58\decr{74} \\
LLaVA & APGD & 87.51 & 52.06\decr{41} & 8.11\decr{91} \\
LLaVA & CW & 87.51 & 80.64\decr{8} & 77.1\decr{12} \\
BLIP2-T5 & PGD & 86.47 & 28.64\decr{67} & 2.98\decr{97} \\
BLIP2-T5 & APGD & 86.47 & 31.39\decr{67} & 2.37\decr{97} \\
BLIP2-T5 & CW & 86.47 & 70.11\decr{19} & 58.85\decr{32} \\
InstructBLIP & PGD & 89.89 & 21.09\decr{77} & 3.66\decr{96} \\
InstructBLIP & APGD & 89.89 & 22.35\decr{75} & 2.18\decr{98} \\
InstructBLIP & CW & 89.89 & 37.81\decr{58} & 31.91\decr{64} \\
\midrule
 \multicolumn{5}{c}{LMM with Context Acc (\%)} \\
LLaVA  & PGD & 93.74 & 73.62\decr{21} & 57.06\decr{39} \\
LLaVA & APGD & 93.74 & 72.61\decr{23} & 37.65\decr{60} \\
LLaVA & CW & 93.74 & 91.76\decr{2} & 90.2\decr{4} \\
BLIP2-T5 & PGD & 97.67 & 87.54\decr{10} & 94.92\decr{3} \\
BLIP2-T5 & APGD & 97.67 & 87.29\decr{11} & 98.43\incr{1} \\
BLIP2-T5 & CW & 97.67 & 94.97\decr{3} & 92.76\decr{5} \\
InstructBLIP & PGD &  88.94& 66.92\decr{25} &  71.61\decr{19}\\
InstructBLIP & APGD &  88.94&  68.74\decr{23} & 89.22\decr{0} \\
InstructBLIP & CW &  88.94 &  84.92\decr{5} &  82.51\decr{7}\\
 \bottomrule
\end{tabular}
\caption{Top-1 image classification result on COCO 2014val. The first table section shows visual encoder accuracy, referring to CLIP/EVA-CLIP's accuracy on classification; second section shows LMMs' accuracy; third section show LMMs' accuracy, after the context is added to the query. Numbers in parenthesis show \% change w.r.t. the Pre-attack accuracy.}
\label{coco_classification}
\end{table}

\begin{table*}
\small
\centering
\begin{tabular}{llccccccccc}
\toprule
 & &  \multicolumn{3}{c}{LMM Query Decomp. Acc(\%)} & \multicolumn{3}{c}{LMM Plain Acc (\%)} & \multicolumn{3}{c}{Visual Encoder Acc (\%)} \\
   Dataset & Attack & Pre & $\text{Post}_\text{N}$ & $\text{Post}_\text{S}$ & Pre & $\text{Post}_\text{N}$ & $\text{Post}_\text{S}$ & Pre & $\text{Post}_\text{N}$ & $\text{Post}_\text{S}$  \\
   \cmidrule(lr){1-2} \cmidrule(lr){3-5} \cmidrule(lr){6-8} \cmidrule(lr){9-11}
    COCO & PGD & 98.42 & 65.30 \decr{34} & 35.88 \decr{64} & 87.51 & 48.25 \decr{45} & 22.58 \decr{74} & 89.21 & 31.00 \decr{65} & 6.56 \decr{93}\\
    COCO & APGD & 98.42 & 66.98 \decr{32} & 23.88 \decr{76} & 87.51 & 52.06 \decr{41} & 8.11 \decr{91} & 89.21 & 30.01\decr{66} & 0.15 \decr{99}\\
    COCO & CW & 98.42 & 95.27 \decr{3} & 93.68 \decr{5} & 87.51 & 80.64 \decr{8} & 77.10 \decr{12} & 89.21 & 44.92 \decr{50}& 0.04 \decr{99}\\
  Imagenet & PGD & 90.62 & 58.52 \decr{35} & 29.96 \decr{67} & 28.10 & 10.34 \decr{63} & 3.44 \decr{88} & 71.47 & 15.94 \decr{78} & 1.07 \decr{98}\\
  Imagenet & APGD & 90.62 & 57.26 \decr{37} & 27.16 \decr{70} & 28.10 & 11.46 \decr{59} & 4.72 \decr{83} & 71.47 & 4.21 \decr{94} & 0.01 \decr{99}\\
  Imagenet & CW & 90.62 & 87.72 \decr{3} & 86.94 \decr{4} & 28.10 & 21.00 \decr{25} & 19.44 \decr{31} & 71.47 & 11.62 \decr{84} & 0.80 \decr{99}\\
\bottomrule
\end{tabular}
\caption{LLaVA classification accuracy on COCO 2014val and Imagenet 2012val. ``LMM Query Decomp." refers to classification with context and query decomposition, as discussed in Sec.~\ref{subsec: query_decomp}. ``LMM plain" refers to classification without context and query decomposition. ``Visual Encoder" refers to CLIP/EVA-CLIP's classification accuracy. Numbers in parenthesis show \% change w.r.t. the Pre-attack accuracy.}
\label{cls_context_with_multi_qs}
\end{table*}

\subsection{Adding Context Improves LMM Robustness}
To examine the effect of context on LMMs' robustness, we reuse the image classification task. We first ask LLaVA to generate a general one-sentence description for each class. We then insert the generated description corresponding to the correct object into the prompt for querying the LMMs about the main object in the image. Besides the additional context, everything else is kept the same. 

Results are shown in Table~\ref{coco_classification}. We observe that after adding a short sentence of context, the post-attack accuracy for all three LMM models increase by a large margin. In particular, the accuracy drop for BLIP2/InstructBLIP under PGD/APGD reduce to only 20\%, as opposed to an average of 60\% drop without context. Although the resulting accuracy is still not on par with the pre-attack accuracy, it still suggests the efficacy of providing additional context against adversarial input, possibly by helping LLMs recover object attributes from the corrupted visual inputs and match with the correct object. This can be useful when the task is to identify the existence of some target objects (e.g., illicit object detection) from images that could be intentionally manipulated, and in the worst scenario, be adversarial.

Interestingly, we also observe that for BLIP2-T5 and InstructBLIP under PGD and APGD attacks, the $\text{Post}_\text{S}$ accuracy are higher than the normal setting. For APGD, they are even higher than the pre-attack accuracy. We conjecture this is due to the fact that APGD-S is too effective on EVA-CLIP (0.01\% classification accuracy post-attack), and that BLIP2-T5 and InstructBLIP are solely relying on the object description to generate the answer while ignoring the adversarial visual input. The two LLMs therefore hallucinate the object description as the answer. However, although the post-attack accuracy is also low for CLIP (0.15\% under APGD S.), we do not observe the same behavior for LLaVA. Possibly the reason is due to different ways LLaVA and BLIP combine the two modalities. While LLaVA takes visual input as standalone tokens, separately from text tokens, BLIP utilizes a Q-former, which blends two modalities together and therefore possibly outweighing the visual input signal with text's.

\subsection{Towards Real-World Application: Context-Augmented Image Classification}
\label{subsec: query_decomp}

In the previous section, we show that adding the correct object context enhances LMMs' robustness against adversarial images. In practice, the correct context is typically unknown. However, in the case of closed-world image classification, where the list of object classes are fixed, we can decompose each question into multiple existence questions. Each question queries the presence of one object class, along with the context corresponding to that object. Afterwards, we choose the object with the highest confidence from the LLM's final projection head. We term our approach query decomposition.

While this solution may appear to be brute-force, such a scheme is inherently able to support making each query in parallel, thereby improving the efficiency. Nevertheless, we would like to demonstrate how we can apply our findings towards real-world setting, and hope this approach presents a viable starting point that opens the door to future work.
%
%
%
To see whether our proposed query decomposition may work, we conduct experiments using COCO 2014val and Imagenet~\cite{deng2009imagenet} 2012 val. For each image, we randomly select 20 object classes while ensuring the correct object class is included. Results are shown in Table~\ref{cls_context_with_multi_qs}. We again observe noticeable improvements on robustness, just like in Table~\ref{coco_classification}. For example, with the context and query decomposition under ``LMM Query Decomp", the percent drops for COCO are mostly 10\% smaller, and 20\% smaller for Imagenet, comparing to post-attack accuracy drops without context as shown under ``LMM plain Acc" columns.

Notably, when query decomposition is utilized to insert context, LMM's performance on ImageNet classification is greatly boosted. This can be seen by comparing the pre-attack performance under ``LMM Query Decomp" and under ``LMM Plain Acc".

\section{Conclusion}

In this study, we systematically evaluate the susceptibility of LMMs to visual adversarial inputs across a diverse array of tasks and datasets. Our findings suggests LMMs are highly vulnerable to visual adversarial attacks, even when such adversaries are crafted with respect to the visual model alone. On the other hand, we find that LMMs are ``robust" when the query and attack target does not match. Such characteristics indicates that the traditional task-specific adversarial generation techniques are not universally effective against current LMM, and points to the need for further research into new adversarial attack strategies, particularly in the context of zero-shot inference. Finally, we find adding context about the querying object improves LMMs' visual robustness. We therefore propose a strategy to decompose questions into multiple existence questions associated with the corresponding context, which achieved notable improvements in robustness on COCO and Imagenet classification.

{
    \small
    \bibliographystyle{ieeenat_fullname}

}

\clearpage
\setcounter{page}{1}
\maketitlesupplementary

 \begin{figure*}[ht]
    \centering
    \includegraphics[width=\textwidth]{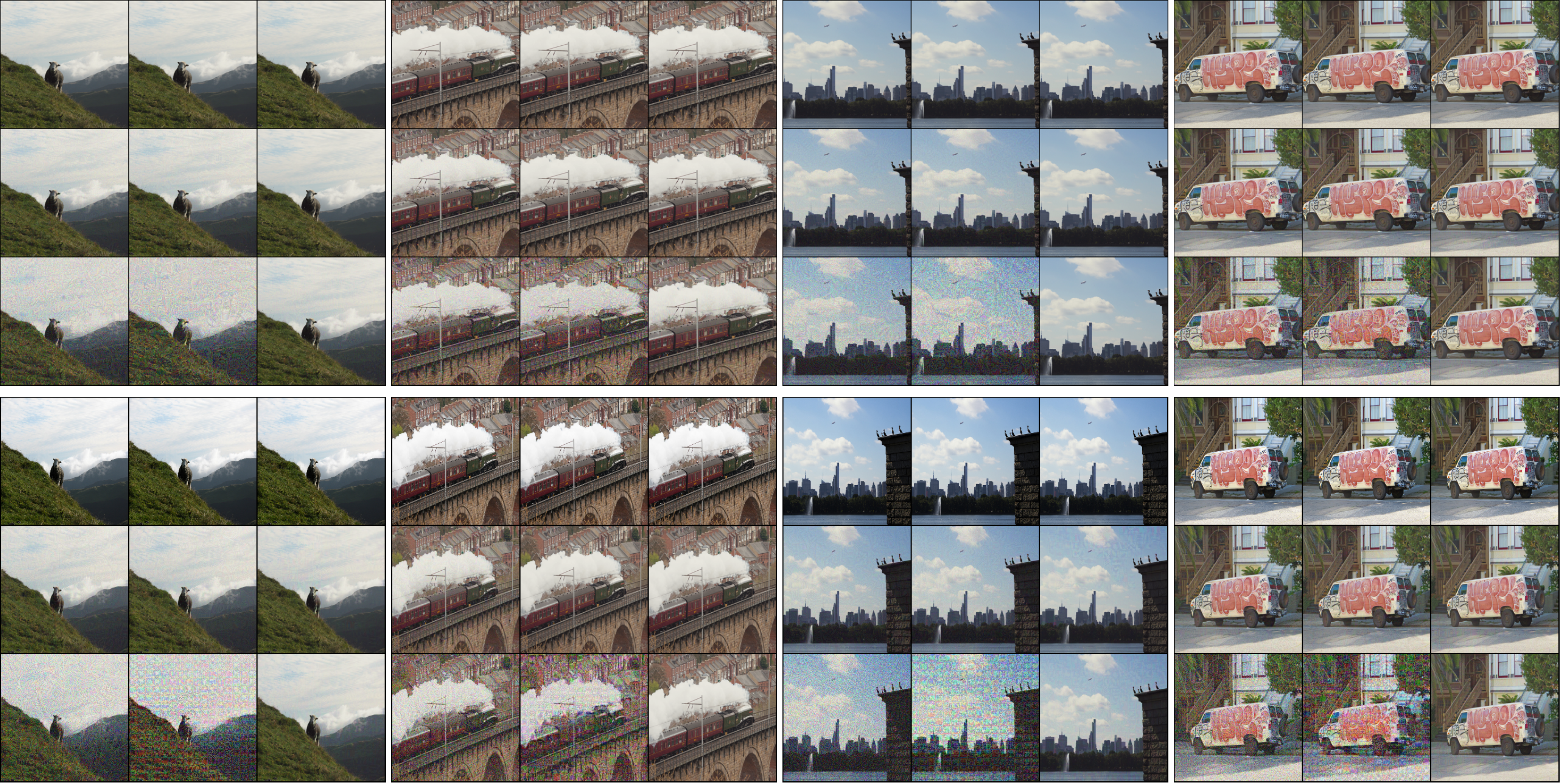}
    \caption{Visualization of three attacks generated from CLIP (top) and EVA-CLIP (bottom). In each of the 3 $\times$ 3 cell, top/mid/bottom row is from clean/normal/strong, and left/mid/right column is from PGD/APGD/CW attack, respectively. Image source: COCO val2014~\cite{lin2014microsoftcoco}.}
    \label{fig:adv_samples}
\end{figure*}

\section{Attacks}

We use implementations from torchattacks~\cite{kim2020torchattacks} to generate adversarial images. All the attacks are un-targeted. Below in Fig.~\ref{fig:adv_samples}, we show more visualizations of adversarial images under different attacks, attack strength and models. 

Among three types of attacks, APGD~\cite{pmlr-v119-croce20b} generates the most perceptible perturbations, as can be seen in Fig.~\ref{fig:adv_samples}. On the contrary, CW yields the most imperceptible perturbations, even under the strong setting. All the three attacks are imperceptible under the normal setting.

By comparing adversarial images generated by CLIP and EVA-CLIP (top and bottom row from Fig.~\ref{fig:adv_samples}), we can observe adversarial perturbations generated under EVA-CLIP are generally more perceptible than that of CLIP's. We can also observe the interestingly highlighted patch border for EVA-CLIP under $\text{APGD}_{\text{S}}$, which does not exist on that of CLIP's.

\section{VQA Per Question-type Accuracy Drop} 
\begin{figure*}[ht]
    \centering
    \includegraphics[width=\textwidth]{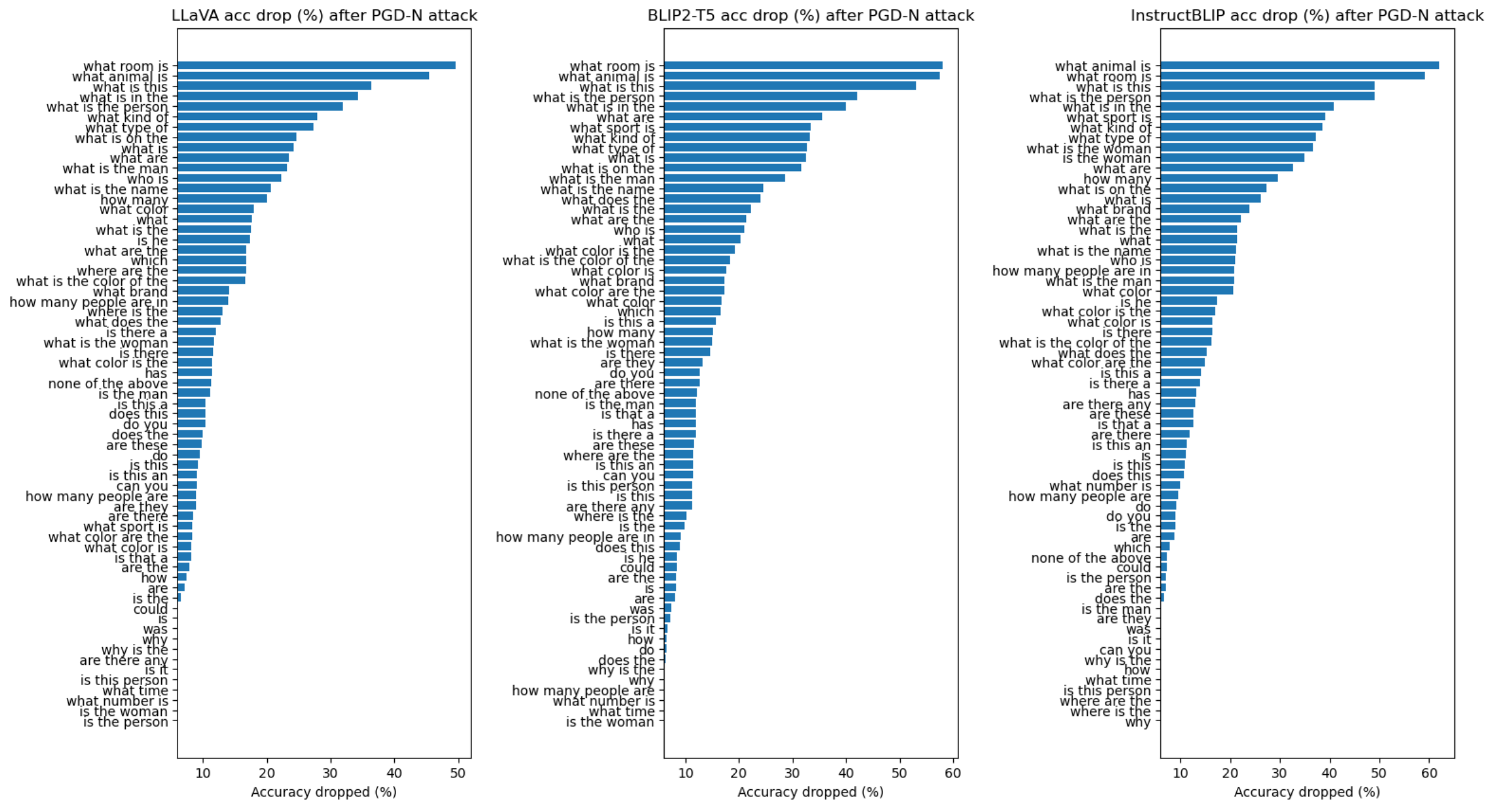}
    \caption{VQA V2 per question-type accuracy drop on LLaVA, BLIP2-T5 and InstructBLIP under $\text{PGD}_{\text{N}}$.}
    \label{fig:acc_drops}
\end{figure*}

Fig.~\ref{fig:acc_drops} shows more results on per-question accuracy drop after adversarial attack. We observe consistent behavior across three tested LMMs (LLaVA~\cite{liu2023visual}, BLIP2-T5~\cite{li2023blip2} and InstructBLIP~\cite{dai2023instructblip}, where accuracy drop the most on questions querying object types or attributes, such as ``what animal/room/kind/type...". 

\section{LLM Responses to Adversarial Visual Questions} 

\begin{figure*}[ht]
    \centering
    \includegraphics[width=\textwidth]{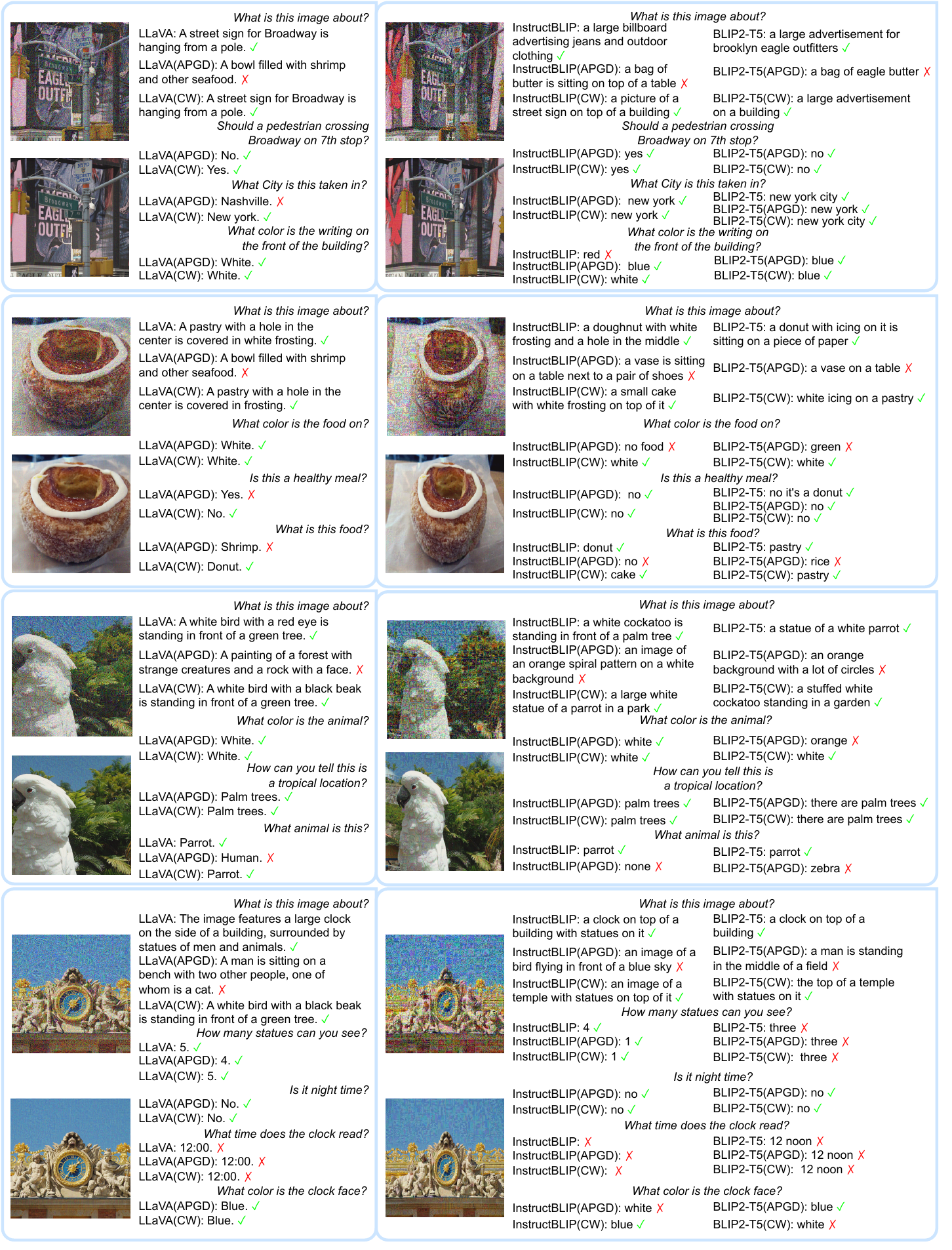}
    \caption{A comparison between the response from three LMMs (LLaVA, BLIP2-T5 and InstructBLIP) on adversarial image generated by $\text{APGD}_{\text{S}}$ and $\text{CW}_{\text{S}}$ with CLIP for LLaVA (left), and EVA-CLIP for BLIP2-T5 and InstructBLIP (mid and right). ``(adv)" refers to LMM's response with the adversarial image. Within each cell, the top/bottom adversarial image is generated by $\text{APGD}_{\text{S}}$/$\text{CW}_{\text{S}}$, respectively. We show the clean response when the adversarial response is different. Image and questions source: VQA V2~\cite{goyal2017making}.}
    \label{fig:llm_responses}
\end{figure*}

\begin{figure*}[ht]
    \centering
    \includegraphics[width=\textwidth]{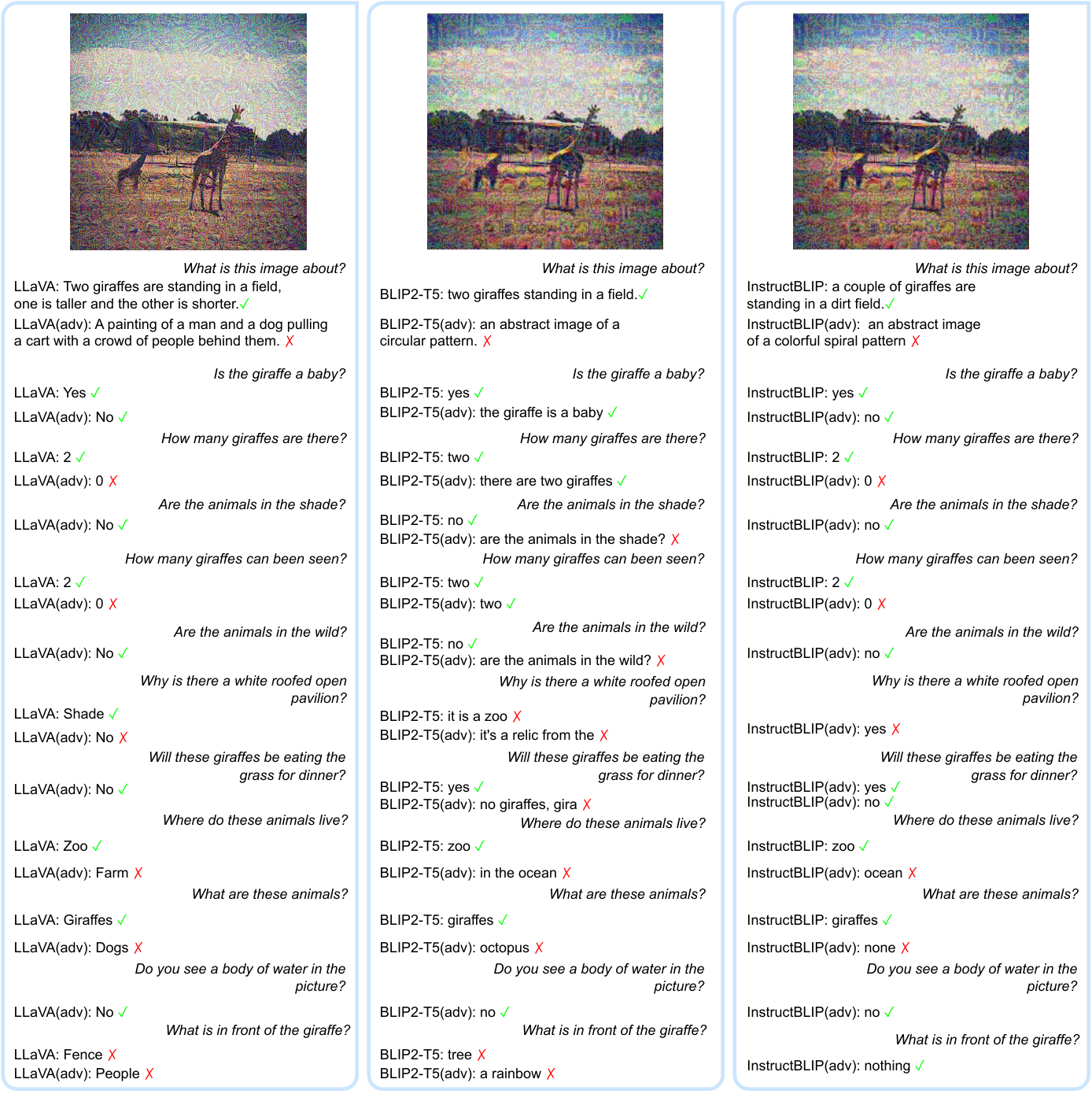}
    \caption{A comparison between responses from three LMMs (LLaVA, BLIP2-T5 and InstructBLIP) on adversarial images generated by $\text{APGD}_{\text{S}}$ with CLIP for LLaVA (left), and EVA-CLIP for BLIP2-T5 and InstructBLIP (mid and right). ``(adv)" refers to LMM's response with the adversarial image. We show the clean response when the adversarial response is different. Image and questions source: VQA V2~\cite{goyal2017making}.}
    \label{fig:llava_response_giraff}
\end{figure*}

In Fig.~\ref{fig:llm_responses} and~\ref{fig:llava_response_giraff}, we show more visualization on the three evaluated LMMs' responses to APGD and CW attacks under the strong setting. We can again observe that adversarial images under APGD attack cause all three LMMs to output completely incorrect image descriptions, yet still having correct answers for ``peripheral" questions, especially those querying the backgrounds. Notably, adversarial images generated under CW attack have little impact on all three LMMs, despite the relatively low classification accuracy after CW attack for CLIP and EVA-CLIP.

\end{document}